\documentclass[journal,twoside,web]{ieeecolor}
\usepackage{tmi}
\usepackage{cite}
\usepackage{amsmath,amssymb,amsfonts}
\usepackage{graphicx}
\usepackage{textcomp}
\usepackage{hyperref}
\usepackage{pifont}
\usepackage{soul}
\usepackage{dsfont}
\usepackage{caption}
\usepackage{subcaption}
\usepackage{booktabs}
\usepackage{wrapfig}
\usepackage[normalem]{ulem}
\usepackage{algpseudocode}
\usepackage{algcompatible}
\usepackage{multirow}
\usepackage{booktabs}
\usepackage{makecell}
\usepackage{xcolor, colortbl}
\usepackage{booktabs} 
\usepackage{longtable, array, booktabs}
\usepackage{makecell}
\usepackage{arydshln}
\usepackage{multirow}
\newcolumntype{?}{!{\vrule width 1pt}}
\newcolumntype{|}{!{\vrule width .5pt}}
\definecolor{LightCyan}{rgb}{0.88,1,1}
\definecolor{darkcyan}{rgb}{0,.79,.75}
\def\BibTeX{{\rm B\kern-.05em{\sc i\kern-.025em b}\kern-.08em
    T\kern-.1667em\lower.7ex\hbox{E}\kern-.125emX}}
\definecolor{darkred}{rgb}{0.6148, 0., 0.}
\definecolor{moredarkred}{rgb}{0.5, 0., 0.}
\definecolor{darkgreen}{rgb}{0., 0.6148, 0.}
\definecolor{darkblue}{rgb}{0., 0., 0.6148}
\definecolor{darkgold}{rgb}{0.9, 0.7, 0.}
\definecolor{darkmagenta}{rgb}{.5, 0, .5}
\definecolor{lightyellow}{rgb}{1., 1., 0.95}
\definecolor{lightgrape}{rgb}{0.93, 0.89, 0.98}
\definecolor{steelblue}{rgb}{0.2745,0.5098,0.7059}
\definecolor{lightcrimson}{rgb}{0.9176,0.4471,0.5412}
\definecolor{LightCyan}{rgb}{0.88,1,1}
\definecolor{DarkCyan}{rgb}{0,.79,.75}
\definecolor{LightGray}{rgb}{0.9, 0.9, 0.9}
\definecolor{cvprblue}{rgb}{0.21,0.49,0.74}
\definecolor{segrow}{RGB}{245,247,250} 
\definecolor{detrow}{RGB}{252,245,235} 

\newcommand{\segtoken}{\textcolor{purple!60}{\texttt{\textbf{[seg]}}}}
\newcommand{\dettoken}{\textcolor{blue!60}{\texttt{\textbf{[det]}}}}
\newcommand{\closertoken}{\textcolor{green!60!black}{\texttt{\textbf{[closer]}}}}

\begin{document}
\title{Segmentation, Detection and Explanation: \\ A Unified Framework for CT Appearance Reasoning}
\author{
\parbox{0.98\textwidth}{\centering
Yuyuan~Liu, Can~Peng, Yingyu~Yang, Qianye~Yang, Cheng~Ouyang, and J.~Alison~Noble
}
\thanks{
\noindent Yuyuan Liu and Can Peng contributed equally to this work. 
Cheng Ouyang and J. Alison Noble jointly supervised this project. 
Corresponding author: Yuyuan Liu (e-mail: yuyuan.liu@eng.ox.ac.uk).\\
All authors are with the Institute of Biomedical Engineering, Department of Engineering Science, University of Oxford. 
This work is supported by UKRI grant EP/X040186/1 (Turing AI Fellowship) and UKRI SFTC AIRR Early Access Project ANON-BYYG-VX4C-Z \textit{Machine Learning based Medical Image Analysis Models that Reason with Human Language}.
}
\vspace{-20pt}
}

\maketitle

\begin{abstract}
Recent progress in deep learning has significantly advanced CT image analysis, particularly for
segmentation
tasks. However, these advances are largely confined to image-level pattern recognition, with most methods lacking explicit anatomical or contextual reasoning. 
Large vision–language models introduce linguistic context into image analysis, yet most approaches typically focus on a single task, which is insufficient for clinical workflow analysis that requires multiple fine-grained types of analysis, such as anatomy detection and segmentation.
In this paper, we propose a unified autoregressive framework that integrates 
language-guided visual reasoning
into CT interpretation. Our method introduces task-routing tokens that trigger detection and segmentation heads conditioned on the hidden states of a large vision–language model, enabling coherent generation of 
visual outputs (e.g., masks and bounding boxes)
and textual reasonings. To progressively enhance localisation accuracy and semantic clarity, we further design a “closer-look” mechanism that allows the model to perform progressive coarse-to-fine visits to regions of interest under refined fields of view.
To support model training and evaluation, we curated a new multimodal CT dataset containing pixel-wise masks, bounding boxes, spatial prompts, and structured descriptions for visual objects constructed 
through an AI-assisted annotation process with human verification.
Experiments on public benchmarks demonstrate consistent improvements over the SoTA, achieving up to $1.0\%$ Dice on BTCV and $1.7\%$ Dice on MosMed+, while additionally providing appearance reasoning outputs. The code and dataset will be available.
\end{abstract}


\section{Introduction}
\label{sec:introduction}
\IEEEPARstart{M}{ost} current approaches for medical image analysis operate primarily on visual cues only, often producing suboptimal outcomes for visually challenging cases, such as small structures or anatomically complex regions, typically due to insufficient understanding of an object appearance. We hypothesise that incorporating language signals into the learning process provides a natural way to supplement semantic information of appearance into CT image analysis models. By linking visual patterns with anatomical semantics, language allows a model to capture structure and context explicitly beyond (graylevel) visual pattern matching. 

Autoregressive Large Vision–Language Models (LVLMs) provide a natural framework for jointly modeling visual perception tasks, such as segmentation and detection, with language-guided appearance reasoning. Here, \emph{appearance reasoning} denotes the use of language to explicitly describe and constrain object-level visual characteristics, such as shape, size, and anatomical context.
Recent work~\cite{lai2024lisa, rasheed2024glamm, kyung2025medregion} has introduced task-specific tokens (e.g., \texttt{\small [SEG]}) to produce segmentation masks, enabling tighter coupling between language and visual predictions.
However, focusing on a single task such as segmentation is insufficient for real-world CT diagnostic settings~\cite{park2024comprehensive, elhanashi2025ai, cai2025systematic}. 
Clinical workflows can require multiple tasks to be completed depending on the clinical objective. For instance, for small or ambiguous structures object detection~\cite{lin2017focal, jaeger2020retina} may localise small regions, while segmentation is well-suited for detailed delineation during downstream assessment~\cite{li2025towards}, such as reporting~\cite{liu2025automatic}, or treatment planning~\cite{ding2025ai}. 
These considerations suggest the need for a unified, multimodal reasoning framework to support segmentation, detection, enhanced by appearance reasoning. Such a framework should allow models to flexibly switch between tasks while maintaining coherent visual perception with reasoning. \\
\indent Beyond producing different outputs, clinical interpretation typically follows a structured (often \textit{sequential}) process in which decisions are progressively refined as additional image-based evidence is examined~\cite{siviengphanom2024computer, gandomkar2019visual}.  Typically, clinicians first review the CT slice as a whole to establish a global anatomical context, and then selectively focus on regions that appear suspicious or ambiguous~\cite{gandomkar2023reliability, kundel1975interpreting, kundel2007holistic}.
Despite this being a fundamental principle of human visual perception, most existing AI systems perform the task in a single forward pass, without revisiting ambiguous regions using local context and evidence acquired in previous passes.
\\
\indent In this paper, we propose a unified autoregressive framework for CT image interpretation that integrates fine-grained image interpretation such as object detection and semantic segmentation with language-based object appearance reasoning. To this end the framework employs task-routing tokens to invoke computational tasks, including segmentation, detection, and a controlled region-focused refinement step based on \textit{prior evidence} acquired in previous analysis passes.
Specifically, the model first performs an analysis of the entire CT slice to establish global anatomical context. When further inspection is required, a dedicated refinement token triggers a \textit{closer-look} operation, in which a region of interest is selected and re-examined.
This two-stage visual perception and reasoning process enables the model to balance global context with local visual evidence. Notably, to enhance analysis performance on small objects, the workflow incorporates appearance reasoning. It utilises detailed, input-specific textual descriptions of object appearances as auxiliary supervision signals. By requiring the model to articulate image-specific object appearances a deeper structural understanding is produced. Empirically, this approach is shown to yield  superior performance relative to models trained with only image-based supervision.\\
\indent To support the training and evaluation of this framework, we curated a multimodal abdominal CT dataset, built upon the 
widely used \cite{landman2015miccai}, which we call \textit{BTCV++}. \textit{BTCV++} is intended as a dataset for abdominal CT radiological interpretation. It has structured vision–language annotations, namely
each organ instance within a CT slice is associated with pixel-level segmentation masks, bounding boxes, and descriptive text that capture anatomical context, including shape, size, and spatial relationships with surrounding organs.
Since radiologists typically interpret CT scans in a slice-wise manner rather than a 3D volume at once, a slice-level benchmark naturally reflects how an assistive-AI system could work with a clinician in a real-world diagnostic workflow.

In summary, this paper makes three main contributions:
\begin{enumerate}
\item we propose a unified autoregressive framework for CT interpretation that jointly performs segmentation and detection with appearance reasoning;
\item One notable feature of the design is an iterative image-based reasoning scheme for detection and segmentation, enabling the model to revisit regions of interest and refine predictions for more accurate localisation and delineation; and
\item we curate a new image-text dataset, \textit{BTCV++}, with multi-task annotations: segmentation masks, bounding boxes, and structured appearance descriptions.
\end{enumerate}
We demonstrate that our approach achieves state-of-the-art performance in both segmentation and detection on the \textit{BTCV++} dataset and two public datasets (vanilla BTCV~\cite{landman2015miccai} and MosMed+~\cite{morozov2020mosmeddata}). 

\section{Related Work}

\subsection{CLIP-based Vision–Language Grounding}
CLIP-based vision-language models align image representations with textual descriptions, and enable integration of linguistic cues such as organ names or descriptive phrases into medical image analysis~\cite{li2020comparison, chen2024survey, lai2024bridging, zhang2024mediclip, liu2023clip}.
By learning a shared vision-language embedding space, these models introduce semantic priors at the representation level and support flexible learning paradigms, including partial supervision, few-shot adaptation, and zero-shot generalisation~\cite{liu2023clip, lai2024bridging, koleilat2024medclip, yu2024cp}.
However, the role of language in most existing CLIP-based frameworks is primarily limited to high-level supervision, such as radiological reports or label categories \cite{chen2024causalclipseg, liu2023clip, koleilat2024medclip}.
Additionally, the contribution of language in CLIP-based fine-grained analysis is questionable in some applications as during supervised training the imaging modality can provide sufficient signal to model the task.
From a network architecture perspective, CLIP itself does not offer language or a multi-modal decoding capability, which limits its flexibility in achieving advanced functions such as language outputs (\textit{e.g.,} reasoning, tool-calling commands, \textit{etc}).

\subsection{Autoregressive Vision-Language Grounding}
\noindent Recent studies have explored Large Vision Language Models (LVLMs) that unify visual perception and reasoning within a generative framework~\cite{lai2024lisa, rasheed2024glamm}.
LVLMs are autoregressive models that generate text or structured outputs token by token, conditioned on paired image–text inputs.
This design enables language-enhanced perception across complex anatomical and pathological variations \cite{lai2024lisa}.
Unlike traditional alignment-based VLMs (\textit{i.e., the CLIP family})~\cite{radford2021learning, li2021align}, autoregressive LVLMs can generate free-text language outputs, such as natural language reasoning that support model decisions. 

\noindent In the natural image domain, autoregressive frameworks have become increasingly mainstream, powering applications that unify visual perception and cognition.
Recent approaches \cite{lai2024lisa, zhang2024mm} enhance these models by integrating instruction-following capabilities into autoregressive architectures.
This combination merges human-aligned prompting with structured, token-level generation, allowing models to perform vision–language tasks through natural-language control.
Supervised fine-tuning (SFT) is commonly employed within instruction-following frameworks to strengthen the fine-grained visual understanding capability of LVLMs.
A typical strategy in SFT \cite{lai2024lisa, xia2024gsva, rasheed2024glamm} is to enable LVLMs to emit special tokens (e.g., \texttt{[SEG]}) that trigger pretrained segmentation models such as SAM \cite{ravi2024sam} or SAM2 \cite{kirillov2023segment} to generate query-aligned masks.
This embedding-as-mask capability has also been extended to, for instance, object detection \cite{wang2024visionllmv2} and fine-grained segmentation \cite{lai2024lisa, xia2024gsva, rasheed2024glamm}. 
The extension of autoregressive and instruction-following approaches to medical imaging has shown promising results \cite{llava_med, pmc_llava}.
Recent efforts have applied LVLMs to CT data for free-form report generation \cite{chen2025mimo, bai2024m3d, hamamci2024ct2rep, li2025towards, kyung2025medregion}.
However, applying autoregressive LVLMs to CT data remains challenging due to the increased complexity of anatomical localisation, segmentation, and interpretability requirements.
In real-world diagnostic workflows, clinicians often require multiple structured outputs within a single system:
first detection for small structures or subtle lesions and subsequent fine-grained segmentation for quantitative assessment and treatment planning \cite{hosny2018artificial, borys2023explainable}. 
In this paper, we focus on slice-wise CT visual perception as a primary task and propose a unified autoregressive LVLM that is able to simultaneously perform segmentation and detection, supported by natural language reasoning, within a single framework.

\section{Methodology}
\label{subsec:Method}
\begin{figure*}[t]
  \centering
  \includegraphics[width=.95\linewidth]{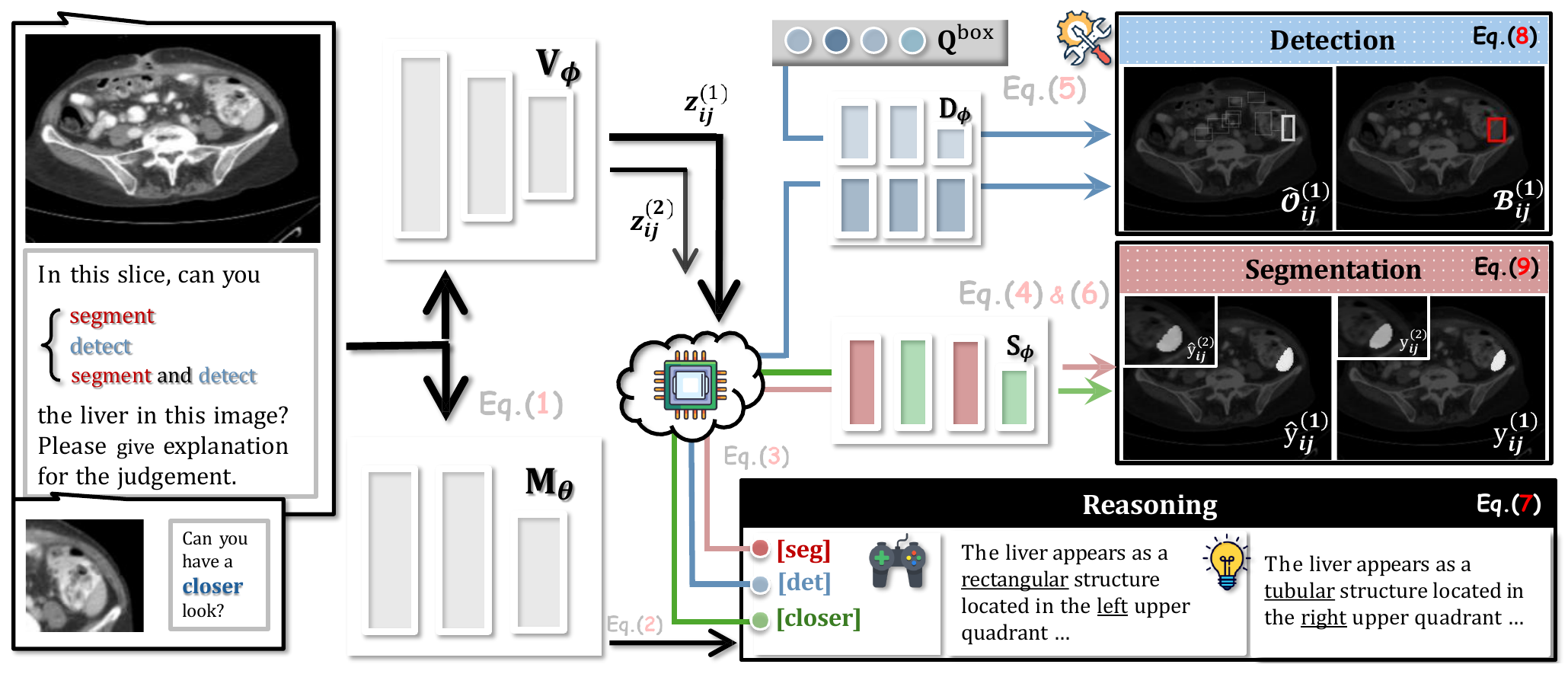}
  \vspace{-5pt}
\caption{\textbf{Illustration of our approach.} 
Given a CT slice $x_i$ and a language query $q_{ij}$ for a visual entity (\textit{e.g.,} an anatomical structure), the LVLM generates a response $\hat{a}_{ij}^{(1)}$ together with routing tokens (\textit{e.g.,} \segtoken, \dettoken), which activate the segmentation and detection branches to produce visual predictions $\hat{y}_{ij}^{(1)}$ in Eq.~\eqref{eq:seg} and \(\hat{\mathcal{O}}_{ij}^{(1)}\) in Eq.~\eqref{eq:det}. The visual outputs are supervised by the perception loss from Eq.~\eqref{eq:l_seg} and Eq.~\eqref{eq:l_det}, while the textual response is supervised by the reasoning loss from \eqref{eq:l_language}. When fine-grained inspection is required, a refinement token \closertoken\ from Eq.~\eqref{eq:close} triggers a region-focused analysis, where the model focuses on a zoomed-in patch $\tilde{x}_{ij}$ to obtain a refined segmentation mask $\hat{y}_{ij}^{(2)}$.
}
  \label{fig:framework}
  \vspace{-10pt}
\end{figure*}
\noindent 
\noindent \textbf{Framework overview.} We consider a dataset 
\(\mathcal{D} = \{(x_i, o_i)\}_{i=1}^{|\mathcal{D}|}\),
where \(x_i \in \mathbb{R}^{H \times W}\) denotes the \(i\)-th CT slice and
\(o_i = \{(q_{ij}, a_{ij}, y_{ij}, b_{ij})\}_{j=1}^{J_i}\) represents a set of
organ-level annotations associated with \(x_i\).
For each organ instance \(j\), \((q_{ij}, a_{ij})\) is a natural language
question–answer pair, \(y_{ij} \in \{0,1\}^{H \times W}\) is a binary
segmentation mask aligned with the input spatial resolution, and
\(b_{ij} \in \mathbb{R}^{K_{ij} \times 4}\) denotes the corresponding bounding
boxes, where \(K_{ij}\) is the number of box annotations for that organ in the
slice.

As shown in Fig. \ref{fig:framework}, our framework is defined as \(\mathcal{F} = \{\mathbf{M}_\theta, \mathbf{V}_\phi\}\),
where \(\mathbf{M}_\theta\) is a large vision–language model (LVLM) for
multimodal reasoning, and \(\mathbf{V}_\phi\) is a visual perception model for fine-grained spatial
prediction.
Specifically, the it is formulated as
\(\mathbf{V}_\phi = \mathbf{E}_\phi \circ \{\mathbf{D}_\phi \oplus \mathbf{S}_\phi\}\),
where \(\mathbf{E}_\phi\) denotes a shared visual encoder, and
\(\mathbf{D}_\phi\) and \(\mathbf{S}_\phi\) are the detection and segmentation
branches.
We introduce a set of learnable box query embeddings
\(\mathbf{Q}^{\mathrm{box}} = \{\mathbf{q}^{\mathrm{box}}_{k}\}_{k=1}^{Q}\),
where \(\mathbf{q}^{\mathrm{box}}_{k} \in \mathbb{R}^{d}\) and \(Q\) is the number of box queries.
These embeddings are shared across images and learned jointly with the model.
To couple linguistic reasoning with visual perception, we introduce
task-specific routing tokens that are generated by the LVLM during decoding.
In particular, the \segtoken\ and \dettoken\ tokens are used to activate and
condition the segmentation and detection branches, respectively.
In addition, a dedicated \closertoken\ token enables an optional second-stage
forward pass, allowing the model to perform region-focused refinement for
fine-grained visual prediction.
\subsection{CT image Interpretation with Appearance Reasoning}
\noindent As illustrated in Fig.~\ref{fig:framework}, our framework incorporates  both image analysis tasks (incl. segmentation and detection) and the generation of language reasoning. \\
\textbf{Appearance reasoning}. Given an image–query pair \((x_i, q_{ij})\), we first utilise \(\mathbf{M}_\theta\) to generate an answer sequence \(\hat{a}^{(1)} = (\hat{a}^{(1)}_1, \dots, \hat{a}^{(1)}_T)\), using autoregression, formulated as:
\begin{equation}
\mathbf{M}_\theta\!\left(\hat{a}^{(1)} \mid x_i, q_{ij}\right)
= \prod_{t=1}^{T} \mathbf{M}_\theta\!\left(\hat{a}^{(1)}_t \mid \hat{a}^{(1)}_{<t},\, x_i,\, q_{ij}\right),
\end{equation}
where \(T\) is the total number of output tokens and \(\hat{a}^{(1)}_{<t} = (\hat{a}^{(1)}_1, \dots, \hat{a}^{(1)}_{t-1})\) denotes the previously generated tokens.
When a query requires fine-grain visual reasoning, the language model $\mathbf{M}_\theta$ generates a task-specific routing token corresponding to segmentation or detection. The generation of routing tokens is learned during training under task-level supervision, allowing the model to autonomously decide whether and how a visual perception module should be called.
Formally, the token at step $t$ is defined as:
\begin{equation}
\resizebox{.99\hsize}{!}{$
\hat{a}^{(1)}_t =
\begin{cases}
\segtoken, & \text{if the intent is segmentation}, \\[2pt]
\dettoken, & \text{if the intent is detection}, \\[2pt]
\text{a standard text token}, & \text{otherwise.}
\end{cases}$}
\end{equation}
\textbf{Task-semantics joint embedding.} To bridge the reasoning module with the visual perception module, we extract the hidden representations of the routing tokens from the last hidden layer of \(\mathbf{M}_\theta\):
\begin{equation}
\begin{aligned}
\mathbf{e}_{ij}^{\segtoken} &= \mathrm{HiddenState}(\hat{a}_{ij}, \segtoken), \\
\mathbf{e}_{ij}^{\dettoken} &= \mathrm{HiddenState}(\hat{a}_{ij}, \dettoken).
\end{aligned}
\end{equation}
These token embeddings serve as semantic conditioning signals for the corresponding visual branch, guiding them to produce predictions that are grounded on the object appearances, reasoned by the language model. 
Meanwhile, the visual model encodes the input image via the shared encoder, producing a latent feature map \(z_i = \mathbf{E}_\phi(x_i)\). \\
\noindent \textbf{Segmentation.} The visual module produces a pixel-wise mask
conditioned on the segmentation embedding:
\begin{equation}
\hat{y}_{ij}^{(1)} = \mathbf{S}_\phi\!\left(z_i;\, \mathbf{e}_{ij}^{\segtoken}\right),
\label{eq:seg}
\end{equation}
where \(\hat{y}_{ij}^{(1)} \in \{0,1\}^{H \times W}\) denotes the predicted mask. \\
\noindent \textbf{Detection.} The detection branch predicts a set of object hypotheses conditioned on the detection embedding:
\begin{equation}
\hat{\mathcal{O}}_{ij}^{(1)}
=
\left\{
\left(\hat{b}_{ij}^{k},\, \hat{s}_{ij}^{k}\right)
\right\}_{k=1}^{Q}
=
\mathbf{D}_\phi\!\left(z_i;\, \mathbf{e}_{ij}^{\dettoken}, \mathbf{Q}^{\mathrm{box}}\right),
\label{eq:det}
\end{equation}
where \(\hat{b}_{ij}^{k}\) denotes the predicted bounding box coordinates and
\(\hat{s}_{ij}^{k}\) denotes the associated objectness score.
\vspace{-5pt}
\subsection{A Closer Look for Fine-grained Results}
To support fine-grained visual understanding, our system enables a follow-up reasoning step formulated in a dialogue-based manner.
Specifically, when the user explicitly requests more detail (e.g., \texttt{Can you have a closer look?}), the system identifies a region-of-interest (ROI) from the current outputs and re-enters the reasoning loop.
Let \(\tilde{x}_{ij}\) denote the cropped region-of-interest (ROI)  containing the \(j\)th object,
which is derived from the ground-truth segmentation mask \(y_{ij}\).
The cropped ROI \(\tilde{x}_{ij}\) is resized to \(H \times W\) and forms a new
image–query pair \((\tilde{x}_{ij}, q_{ij}^{(2)})\), where \(q_{ij}^{(2)}\) represents
a second-round follow-up query, often conditioned on the dialogue history or
explicitly referencing prior responses.
The model then repeats the multimodal process to receive predicted answer \(\hat{a}^{(2)} = \mathbf{M}_\theta(\tilde{x}_{ij}, q_{ij}^{(2)})\) and the image embedding
\(z_{i}^{(2)} = \mathbf{E}_\phi(\tilde{x}_{ij})\). Finally, the fine-grained segmentation and resized ROI can be observed with:
\begin{equation}
\hat{y}_{ij}^{(2)} = \mathbf{S}_\phi\!\left(z_{ij}^{(2)};\, \mathbf{e}_{ij}^{\closertoken}\right), 
\label{eq:close}
\end{equation}
where \(\mathbf{e}_{ij}^{\closertoken}\) is an updated token embedding. These additional predictions provide a refined reasoning result.

\label{subsec:Training}
\subsection{Training Objectives}

\noindent We incorporate both the initial and refined reasoning results into the training objective by indexing over dialogue rounds. \\
\noindent \textbf{Reasoning Objective.} For each training sample,
we index the dialogue round by \(r \in \{1,2\}\) and supervise the generated response
with a standard token-level cross-entropy:
\begin{equation}
\ell_{\mathrm{Language}}(\mathcal{D},\, \mathbf{M}_\theta) = 
\frac{1}{\sum_{i=1}^{|\mathcal{D}|} J_i} 
\sum_{i=1}^{|\mathcal{D}|} \sum_{j=1}^{J_i} \sum_{r=1}^{2}
\ell_{\mathrm{CE}}(a_{ij}^{(r)}, \hat{a}_{ij}^{(r)}),
\label{eq:l_language}
\end{equation}
where \(a_{ij}^{(r)}\) and \(\hat{a}_{ij}^{(r)}\) denote the ground-truth and predicted token sequences and \(J_i\) is the object number in the slice. \\
\noindent \textbf{Segmentation Objective.} For segmentation, we apply supervision in both rounds.
The segmentation loss is defined as:
\begin{equation}
\ell_{\mathrm{seg}}(\mathcal{D},\, \mathbf{E}_\phi,\, \mathbf{S}_\phi)
=
\frac{1}{2\sum_{i=1}^{|\mathcal{D}|} J_i}
\sum_{i=1}^{|\mathcal{D}|} \sum_{j=1}^{J_i} \sum_{r=1}^{2}
\ell_{\mathrm{seg}}\!\left(y_{ij}^{(r)}, \hat{y}_{ij}^{(r)}\right),
\label{eq:l_seg}
\end{equation}
where \(\ell_\mathrm{seg}\) is a weighted combination of binary cross-entropy
and Dice loss.\\
\noindent \textbf{Detection Objective.}
The model predicts a set of \(Q\) object hypotheses
\(\hat{\mathcal{O}}_{ij}^{(r)}=\{(\hat{b}_{ij}^{k,(r)}, \hat{s}_{ij}^{k,(r)})\}_{k=1}^{Q}\).
Given the ground-truth bounding boxes
\(\mathcal{B}_{ij}^{(r)}=\{b_{ij}^{n,(r)}\}_{n=1}^{N_{ij}^{(r)}}\),
we compute a bipartite match between predictions and ground truth via the
Hungarian algorithm using a cost that combines distance and overlap of bounding boxes.
\begin{figure*}
    \centering
    \includegraphics[width=.95\linewidth]{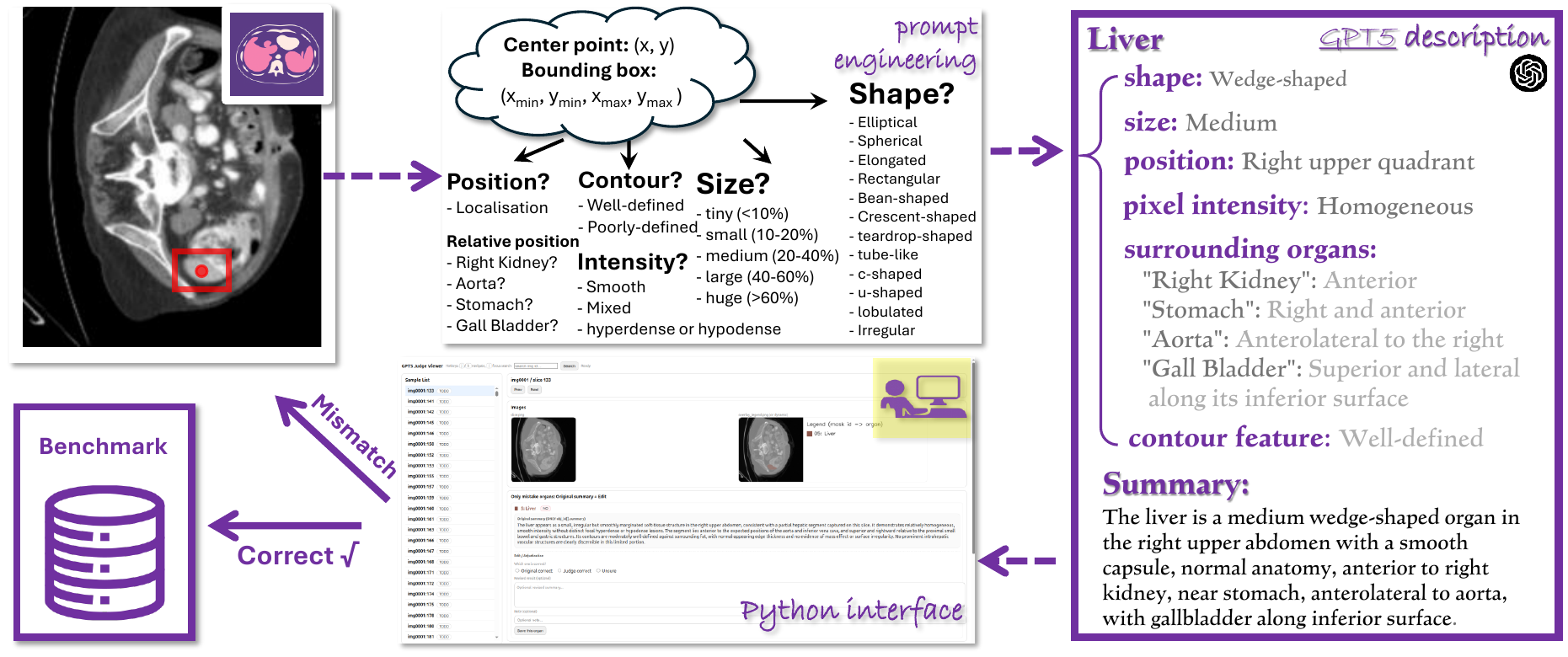}
\caption{\textbf{Human-in-the-loop pipeline for curating appearance description-augmented imaging datasets. } 
Visual prompts localised to objects of interest, including bounding boxes and center points to guide general-purpose LVLM's attention \cite{cai2024vip, shen2025vlm} to enhance the fidelity of generated appearance descriptions. The CT slice, together with appearance descriptions and visual prompts, is sent to LVLM again to generate structured descriptions. All generated appearance descriptions are reviewed and validated through a python-based GUI, with incorrect cases revised or regenerated before inclusion in the dataset.}
    \label{fig:dataset}
\end{figure*}
\begin{figure}
    \centering
    \includegraphics[width=0.49\linewidth]{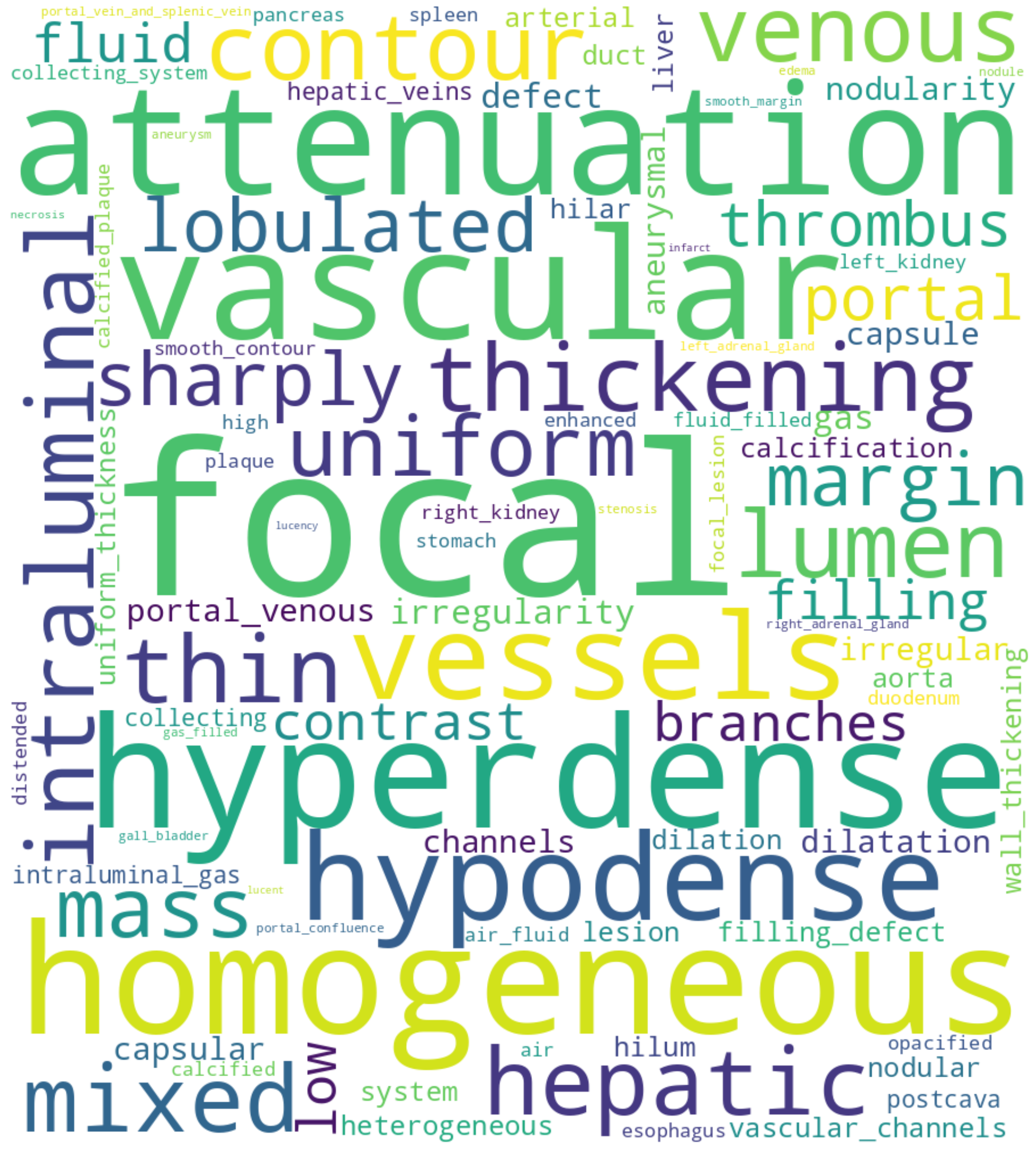}
    \hfill
    \includegraphics[width=0.49\linewidth]{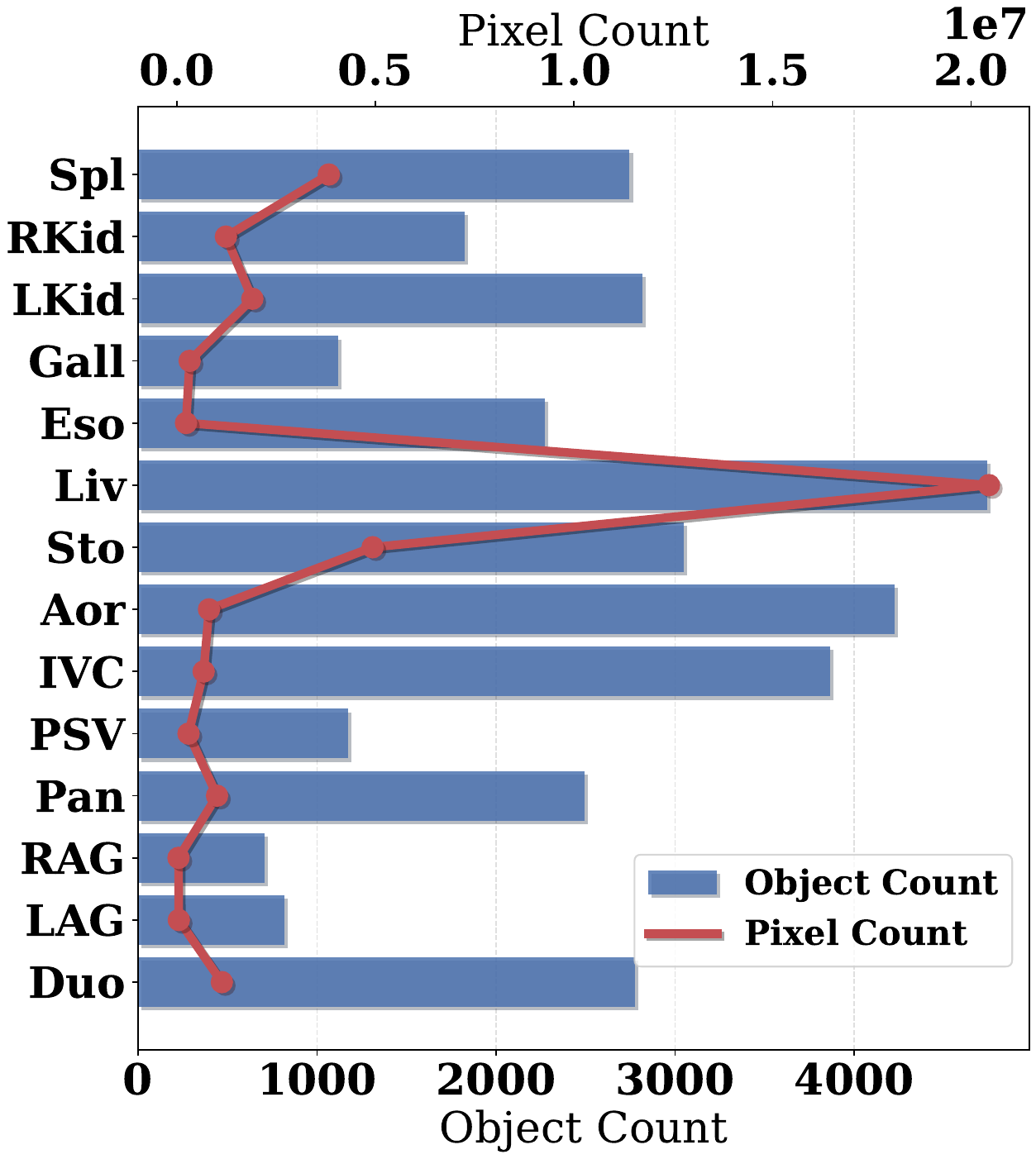}
\caption{\textbf{Textual Description Distribution.} 
Left: Word cloud showing dominant anatomical and descriptive terms in the curated textual corpus. 
Right: Histogram of summary lengths (in words) used for training the reasoning module.}
    \label{fig:data-analys}
    \vspace{-5pt}
\end{figure}
The detection loss consists of two terms:
\begin{equation}
\label{eq:l_det}
\resizebox{.99\hsize}{!}{$
\begin{aligned}
\ell_{\mathrm{det}}(\mathcal{D},\, \mathbf{E}_\phi,\, \mathbf{S}_\phi)
=
\frac{1}{\sum_{i=1}^{|\mathcal{D}|} J_i}
\sum_{i=1}^{|\mathcal{D}|} \sum_{j=1}^{J_i}
\Big[&
\ell_{\mathrm{box}}\!\left(\mathcal{B}_{ij}^{(1)}, \{\hat{b}_{ij}^{k,(1)}\}_{k=1}^{Q}\right)  \\
 & +
\ell_{\mathrm{obj}}\!\left(\mathcal{B}_{ij}^{(1)}, \{\hat{s}_{ij}^{k,(1)}\}_{k=1}^{Q}\right)
\Big],
\end{aligned}$}
\end{equation}
where \(\ell_{\mathrm{box}}\) is defined on matched pairs using L1 and GIoU losses,
and \(\ell_{\mathrm{obj}}\) is a binary cross-entropy loss over objectness scores,
with unmatched queries treated as negatives.
In our two-round setting, detection supervision (\(r=1\)) is typically applied in the first
round, while the second round focuses on fine-grained segmentation.
Finally, the total loss balances reasoning and perception objectives:
\begin{equation}
\begin{aligned}
\mathcal{L}(\mathcal{D},\, \mathbf{M}_\theta,\, \mathbf{E}_\phi,\, \mathbf{S}_\phi,\, \mathbf{D}_\phi)
=\;
& \ell_{\mathrm{Language}}(\mathcal{D},\, \mathbf{M}_\theta) \\
& +\; \lambda_{\mathrm{seg}}\,\ell_{\mathrm{seg}}(\mathcal{D},\, \mathbf{E}_\phi,\, \mathbf{S}_\phi) \\
& +\; \lambda_{\mathrm{det}}\,\ell_{\mathrm{det}}(\mathcal{D},\, \mathbf{E}_\phi,\, \mathbf{D}_\phi),
\end{aligned}
\end{equation}
where the entire training pipeline is performed in an end-to-end manner, where in each iteration, the loss from $\ell_{\text{Language}}(.)$ and $ \ell_{\text{perception}}(.)$ are simultaneously minimised.

\subsection{Inference}
\noindent During inference, given an image–query pair $(x_i, q_{ij})$, the LVLM $\mathbf{M}_\theta$ first generates an appearance reasoning $\hat{a}_{ij}^{(1)}$ together with tokens indicating the requested visual analysis (routing tokens). If a detection token $\dettoken$ is produced, the detection head $\mathbf{D}_\phi$ outputs a set of box hypotheses $\hat{\mathcal{O}}_{ij} = \{(\hat{b}_{ij}^k, \hat{s}_{ij}^k)\}_{k=1}^{Q}$. We select the relevant bounding box coordinates by thresholding objectness scores $\hat{s}_{ij}^k$, which are then returned as detection results or used to guide subsequent region-based reasoning when required.
If a segmentation token $\segtoken$ is generated, the segmentation head $\mathbf{S}_\phi$ predicts a mask $\hat{y}_{ij}$. The segmentation output is directly returned when the query requests pixel-wise delineation. Furthermore, when a global-to-local analysis is triggered (e.g., via $\closertoken$), the framework extracts a ROI $\tilde{x}_{ij}$ from $x_i$ based on $\hat{y}_{ij}$, and resizes it to the canonical resolution, and re-enter the reasoning–perception loop to obtain refined predictions. This unified inference process adaptively switches between reasoning, detection, segmentation, and region-focused refinement without external heuristics.

\section{Dataset Curation Pipeline}
The training data are curated from the BTCV~\cite{landman2015miccai} an abdominal CT public dataset, together with additional expert-annotated 
organ segmentation from publicly available resources~\footnote{\url{https://zenodo.org/records/1169361}}, where the \textit{Duodenum} is further included as an additional anatomical class. To remove redundant axial slices, we perform volume-wise filtering. For each volume, we retain representative slices covering the onset, peak, and offset of each organ, with additional protection for small or complex structures. We discard slices that contain only background or exhibit high foreground mask similarity and negligible area change compared to neighboring retained slices (based on IoU \(\geq\) 0.75). This annotation procedure preserves anatomically informative content while reducing slice redundancy. \\
The overall data curation process follows a two-stage human-in-the-loop pipeline. As shown in Figure~\ref{fig:dataset},
in the first stage, we identify CT slices containing target objects of interest (\textit{e.g.,} organs) and extract their corresponding pixel-wise masks. Based on each mask, we derive visual prompts localized to the target region: specifically,
\begin{itemize}
    \item a bounding box defined by $\{(x_{\min}, y_{\min}), (x_{\max}, y_{\max})\}$;
    \item and the object’s center point $(x_{\text{center}}, y_{\text{center}})$.
\end{itemize}
These visual prompts are incorporated as explicit spatial cues within a predefined structured prompt template, which specifies the input format, the anatomical attributes to be described, and the expected JSON output structure. This design guides a general-purpose LVLM, such as \emph{GPT-5}, to focus on the target structure and ensures consistent, standardised generation of textual descriptions.
In the second stage, each CT slice, together with its visual prompts and structured instructions, is input to a general-purpose LVLM and JSON-formatted appearance descriptions are generated that jointly encode geometric properties (\textit{e.g.,} shape, size, and location) and semantic context (\textit{e.g.,} texture, boundary clarity, and anatomical adjacency), with each anatomical attribute explicitly structured to support downstream parsing and control. The rationale behind this is intuitive: For general-purpose LVLMs, low-level visual description is less dependent on specialized medical domain knowledge.

Finally, all generated descriptions were validated by four human annotators. Incorrect or incomplete outputs were revised or regenerated. This human-in-the-loop process ensured adherence to the defined schema and high data quality. In total, 7{,}742 validated slices textual descriptions were generated.

\section{Experiments}

\begin{table*}[t]
\renewcommand{\arraystretch}{1.3}
\centering
\caption{
Comparison with SoTA methods on BTCV++ for both segmentation and detection tasks.
\colorbox{segrow}{Segmentation results} are reported using Dice, and \colorbox{detrow}{detection results} are reported using mAP@0.1. 
Abbreviations are as follows: Spl: Spleen, RKid/LKid: Right/Left Kidney, Gall: Gall Bladder, Eso: Esophagus, Liv: Liver, Sto: Stomach, Aor: Aorta, IVC: Inferior Vena Cava (Postcava), PSV: Portal \& Splenic Veins, Pan: Pancreas, RAG/LAG: Right/Left Adrenal Gland, Duo: Duodenum.
Best average results are highlighted in bold.}
\vspace{-5pt}
\label{tab:seg_det_unified}
\resizebox{1\linewidth}{!}{
\begin{tabular}{?c?cccccccccccccc?c?}
\specialrule{1.5pt}{0pt}{0pt}

Method
 & Spl & RKid & LKid & Gall & Eso & Liv & Sto & Aor & IVC & PSV & Pan & RAG & LAG & Duo & Mean \\
\cline{1-16}

\rowcolor{segrow}
nnUNet~\cite{isensee2021nnu} &
91.41 &  {87.49} & 85.51 &  {79.11} & 72.89 & 92.68 & 87.91 & 66.83 & 73.81 & 61.88 &  {79.68} & 63.70 & 68.22 & 79.95 & 77.93 \\

\rowcolor{segrow}
TransUNet~\cite{chen2021transunet} &
88.77 & 86.07 & 86.72 & 73.05 & 75.95 & 88.91 & 81.91 &  {89.03} & 80.12 & 67.49 & 74.12 & 60.23 & 58.72 & 71.77 & 77.35 \\

\rowcolor{segrow}
DeepLabV3+~\cite{chen2018encoder} &
91.08 & 81.57 & 87.10 & 70.86 & 74.81 & 90.93 & 85.84 & 79.87 & 75.21 & 65.58 & 75.51 & 58.45 & 56.47 & 72.75 & 76.15 \\

\rowcolor{segrow}
LViT~\cite{li2023lvit} &
90.12 & 83.33 & 88.08 & 73.49 & 69.98 & 90.96 & 84.36 & 88.98 & 77.44 & 61.02 & 71.43 & 57.62 & 61.46 & 72.60 & 76.49 \\

\rowcolor{segrow}
LISA~\cite{lai2024lisa} &
91.41 &  {87.49} & 87.51 &  {79.11} & 80.89 & 92.68 & 87.91 & 66.83 & 73.81 & 62.88 &  {79.68} & 63.70 & 68.22 & 79.95 & 78.72 \\

\cdashline{1-16}
\rowcolor{segrow}
Ours (Janus-1B) &
 {93.39} & 84.67 &  {89.84} & 77.22 & 73.55 &  {95.24} & 82.93 & 85.54 &  {85.79} & 70.60 & 73.63 & 61.16 & 58.58 & 65.38 & 78.39 \\

\rowcolor{segrow}
Ours (Qwen2.5VL-3B) &
91.55 & 87.04 & 89.41 & 77.69 &  {81.58} & 92.55 &  {88.00} & 77.49 & 78.41 &  {72.30} & 79.63 &  {71.55} &  {71.55} &  {80.37} &  \textbf{81.37} \\

\hline

\rowcolor{detrow}
nnDetection~\cite{baumgartner2021nndetection} &
98.31 & 93.77 & 97.33 & 87.65 & 96.66 & 97.80 & 94.15 & 94.31 & 93.29 & 51.76 & 90.89 & 67.29 & 68.80 & 91.28 & 87.38 \\

\rowcolor{detrow}
RetinaUNet~\cite{jaeger2020retina} &
93.24 & 94.57 & 95.20 & 83.35 & 96.73 & 98.16 & 85.09 & 96.40 & 99.72 & 63.10 & 80.30 & 88.79 & 87.54 & 81.80 & 88.86 \\

\rowcolor{detrow}
DETR~\cite{carion2020end} &
98.11 & 98.07 & 97.97 & 94.59 & 97.90 & 98.56 & 95.34 & 97.47 & 97.01 & 65.39 & 93.08 & 77.49 & 85.62 & 93.37 & 92.14 \\

\cdashline{1-16}
\rowcolor{detrow}
Ours (Janus-1B) & 94.93 & 88.95 & 96.16 & 85.96 & 95.22 & 92.63 & 86.18 & 99.11 & 99.80 & 65.04 & 85.25 & 89.61 & 88.26 & 82.88 & 89.28 \\

\rowcolor{detrow}
Ours (Qwen2.5VL-3B) &
100.00 & 100.00 & 100.00 & 82.00 & 100.00 & 97.48 & 98.51 & 100.00 & 100.00 & 65.50 & 79.52 & 91.09 & 91.32 & 85.59 & \textbf{92.21} \\

\specialrule{1.5pt}{0pt}{0pt}
\end{tabular}}
\end{table*}
\begin{table*}[t]
\caption{
Comparison with SoTA methods on BTCV~\cite{landman2015miccai} using five-fold cross-validation.
Results are reported in Dice for all annotated organs.
For fair comparison, model ensemble is not used during evaluation.
Dataset-specific training denotes training on BTCV only,
while dataset-agnostic training denotes training on multiple datasets.
Best results are highlighted in bold.
}
\centering
\scriptsize
\renewcommand{\arraystretch}{1.25}

\resizebox{1\linewidth}{!}{
\begin{tabular}{?c?ccccccccccc?c?}

\specialrule{1.5pt}{0pt}{0pt}
Method &
Spleen &
(L/R) Kidney &
G. Bladder &
Esophagus &
Liver &
Stomach &
Aorta &
IVC &
P. \& S. Vein &
Pancreas &
A. Gland &
Mean \\
\hline
TransUNet~\cite{chen2021transunet}
& 94.10 & 89.53 & 65.49 & 73.19 & 93.24 & 80.85 & 87.47 & 80.48 & 71.47 & 74.26 & 64.76 & 79.16 \\

CoTr~\cite{xie2021cotr}
& 95.51 & 88.61 & 68.49 & 75.83 & 95.93 & 81.84 & 89.01 & 82.32 & 73.39 & 75.12 & 65.78 & 80.48 \\

TransBTS~\cite{isensee2021nnu}
& 94.59 & 89.85 & 68.50 & 75.59 & 96.14 & 83.72 & 88.85 & 82.28 & 74.25 & 75.12 & 66.74 & 80.94 \\

nnFormer~\cite{zhou2021nnformer}
& 94.51 & 90.94 & 65.51 & 74.49 & 96.10 & 83.83 & 88.91 & 80.58 & 75.94 & 77.71 & 68.19 & 81.22 \\

UNETR~\cite{hatamizadeh2022unetr}
& 94.91 & 92.61 & 76.98 & 74.01 & 96.17 & 79.98 & 89.74 & 81.20 & 75.05 & 80.12 & 62.60 & 81.43 \\

nnUNet~\cite{isensee2021nnu}
& \textbf{95.92} & 90.45 & 66.58 & 75.71 & \textbf{96.49} & 86.05 & 88.33 & 82.72 & \textbf{78.31} & 79.17 & 67.99 & 82.01 \\

SwinUNetr~\cite{he2023swinunetr}
& 95.44 & \textbf{93.39} & 77.12 & 74.14 & 96.39 & 80.12 & \textbf{90.02} & 82.93 & 75.08 & \textbf{81.02} & 64.98 & 82.06 \\

\cdashline{1-13}
Ours (Janus1B) 
& 93.98 & 89.79 & \textbf{78.05} & \textbf{76.02} & 93.42 & 85.33 
& 86.59 & \textbf{82.96} & 73.74 & 80.44 & 65.78 & 82.31 \\

Ours (Qwen2.5VL-3B)
& 95.32 & 90.73 & 76.99 & 75.96 & 92.36 & \textbf{86.27} 
& 88.80 & 81.54 & 76.68 & 79.38 & \textbf{69.72} & \textbf{83.07} \\
\specialrule{1.5pt}{0pt}{0pt}
\end{tabular}}
\label{tab:btcv_benchmark_reasoning_style}
\vspace{-10pt}
\end{table*}

\begin{table}[t!]
\centering
\caption{Comparison with SoTA methods on MosMedData+~\cite{landman2015miccai} for the segmentation. Performance is evaluated using Dice and mIoU, with the best results highlighted in bold.}
\label{tab:mosmed+}
\resizebox{0.4\textwidth}{!}{%
\renewcommand{\arraystretch}{1.25}
\begin{tabular}{?l?cc?}
\specialrule{1.5pt}{0pt}{0pt}
\textbf{Method} & \textbf{Dice (\%)} & \textbf{mIoU (\%)} \\
\specialrule{1.5pt}{0pt}{0pt}
nnUNet \cite{isensee2021nnu}         & 72.59 & 60.36 \\
TransUNet \cite{chen2021transunet}  & 71.24 & 58.44 \\
ConVIRT \cite{zhang2022contrastive} & 72.06 & 59.73 \\
TGANet \cite{tomar2022tganet}       & 71.81 & 59.28 \\
GLoRIA \cite{huang2021gloria}       & 72.42 & 60.18 \\
ViLT \cite{kim2021vilt}           & 72.36 & 60.15 \\
LAVT \cite{yang2022lavt}              & 73.29 & 60.41 \\
LViT-T~\cite{li2023lvit}       & 74.57 & 61.33 \\
\rowcolor{black!10}
\cdashline{1-3}
\rowcolor{black!10}
\textbf{Ours (Janus1B)} & 75.46 & 62.39 \\
\rowcolor{black!10}
\textbf{Ours (Qwen2.5VL-3B)} & \textbf{76.23} & \textbf{63.33} \\
\specialrule{1.5pt}{0pt}{0pt}
\end{tabular}}
\vspace{-10pt}
\end{table}

\subsection{Implementation Details}
\noindent \textbf{Network Architecture.}
We employ both Janus-1B~\cite{chen2025janus} and Qwen2.5-VL-3B-Instruct~\cite{bai2025qwen2} as the backbone LVLM. 
The visual module is built on SAM~\cite{kirillov2023segment} with a ViT-H encoder backbone. 
During training, the pretrained weights are loaded for both the image encoder and the segmentation branch.
The latent embeddings of special tokens are projected from the 2048-dimensional LVLM hidden states to the 256-dimensional ViT-H embedding space using a two-layer MLP with GELU activation.
The detection branch consists of two lightweight fully-connected networks, responsible for predicting objectness scores and bounding box coordinates. \\
\noindent \textbf{Training Configuration.}
All modelling software is implemented in PyTorch. Models were built using DeepSpeed distributed training on 4 NVIDIA H100 GPUs. 
Mixed-precision computation with bfloat16 is employed throughout for computational efficiency. 
The input CT images are resized to $1024 \times 1024$, matching the input resolution requirement for the SAM encoder.  
During training, the parameters of the LVLMs are fine-tuned using LoRA adapters applied to the attention projection layers with rank~$r{=}8$, $\alpha{=}16$, and a dropout rate of $0.05$. 
Following~\cite{lai2024lisa}, the image encoder is equipped with visual adapters, whereas the mask decoder is jointly optimized with the semantic projection head and task-specific visual heads.
Each model is trained for 40 epochs using the AdamW optimizer with a learning rate of $3\times10^{-4}$, $(\beta_1, \beta_2) = (0.9, 0.95)$, and no weight decay. 
Gradient accumulation is set to 10 steps, and four worker threads are used for data loading. 
Segmentation supervision combines Dice and binary cross-entropy losses, whereas bounding-box regression employs L1 and generalised IoU losses, all equally. 

\begin{table}[t!]
\renewcommand{\arraystretch}{1.2}
\caption{\textbf{Ablation of closer-look} on the AbdominalReasoningBench dataset. 
We measure the improvements based on Dice and HD95, where the \colorbox{green!10}{green} row results are based on iterative refinement.}
\label{tab:closerlook_ablation}
\centering
\resizebox{.95\linewidth}{!}{
\begin{tabular}{?lcc?}
\specialrule{1.5pt}{0pt}{0pt}
Architecture & Average Dice $\uparrow$ & Average HD95 $\downarrow$ \\
\specialrule{1.5pt}{0pt}{0pt}
w/ Janus1B & 76.83 & 10.13 \\
\rowcolor{green!10}
w/ Janus1B & 78.39 & 9.65 \\
w/ Qwen2.5VL-3B & 80.24 & 6.79 \\
\rowcolor{green!10}
w/ Qwen2.5VL-3B & 81.37 & 5.64 \\
\specialrule{1.5pt}{0pt}{0pt}
\end{tabular}}
\vspace{-5pt}
\end{table}

\begin{table}[t]
\renewcommand{\arraystretch}{1.2}
\caption{\textbf{Ablation study for LoRA rank.} 
Impact of LoRA rank on trainable parameters and segmentation performance based on Qwen2.5VL-3B. Best results are highlighted in bold.}
\label{tab:lora}
\centering
\resizebox{.95\linewidth}{!}{
\begin{tabular}{?cccc?}
\specialrule{1.5pt}{0pt}{0pt}
Rank & Params. (M) & Average Dice $\uparrow$ & HSD $\downarrow$ \\
\specialrule{1.5pt}{0pt}{0pt}
2  & 0.46 & 75.82 & 12.23 \\
4  & 0.92 & 79.15 & 7.16 \\
8  & 1.84 & \textbf{81.37} &\textbf{5.64} \\
16 & 3.69 & 80.92 & 6.63 \\
\specialrule{1.5pt}{0pt}{0pt}
\end{tabular}}
\end{table}

\begin{table}[t]
\centering
\caption{
Qualitative capability comparison on AbdominalReasoningBench based on \texttt{LLM-as-Judge}.
We analyse output reasoning text to different visual and anatomical factors,
including shape, size, abdominal context, and pixel intensity. The evaluation is based on the accuracy.
}
\label{tab:capability_analysis}
\resizebox{\linewidth}{!}{%
\renewcommand{\arraystretch}{1.25}
\begin{tabular}{?l?cccc?}
\specialrule{1.5pt}{0pt}{0pt}
\textbf{Method} 
& \textbf{Shape} 
& \textbf{Size} 
& \textbf{Context} 
& \textbf{Position} \\
\specialrule{1.5pt}{0pt}{0pt}

\textbf{Ours (Janus1B)} & 52.38 & 70.34 & 64.49 & 72.13 \\

\textbf{Ours (Qwen2.5VL-3B)}  & 67.52 & 65.63 & 79.66 & 74.02 \\

\specialrule{1.5pt}{0pt}{0pt}
\end{tabular}}
\end{table}

\begin{table}[t]
\renewcommand{\arraystretch}{1.2}
\centering
\caption{
\textbf{Ablation on challenging organs in detection (mAP@0.1)}.
Mask$\rightarrow$Box extracts boxes via connected components on the threshold;
\emph{area conf.} uses normalised component area as confidence~\cite{wang2024visionreasoner}, while \emph{max-prob conf.} uses the maximum mask probability inside the foreground object.}
\label{tab:ablation_small_organs}
\resizebox{\linewidth}{!}{
\begin{tabular}{?l?cccc?}
\specialrule{1.5pt}{0pt}{0pt}
Method 
& PSV & RAG & LAG & Duo \\
\hline
Detection branch
& \textbf{65.50} & \textbf{91.09} & \textbf{91.32} & \textbf{85.59} \\

Mask$\rightarrow$Box (area conf.)~\cite{wang2024visionreasoner}
& 55.10 & 89.74 & 89.46 & 66.38 \\

Mask$\rightarrow$Box (max-prob conf.)
& 53.13 & 90.07 & 88.22 & 62.46 \\

\specialrule{1.5pt}{0pt}{0pt}
\end{tabular}}
\vspace{-10pt}
\end{table}
\subsection{Datasets}
We utilise the proposed \emph{BTCV++} dataset as described in Section \ref{subsec:Method}. We also use \emph{MosMedData+}~\cite{li2023lvit}, a pulmonary lesion segmentation dataset, to demonstrate that our approach generalises across anatomies (abdominal and thoracic) and types of objects of interest (organs and lesions). 
The vanilla BTCV dataset~\cite{landman2015miccai} is used to evaluate segmentation on abdominal 3D CT volumes. It consists of 30 volumetric abdominal CT scans, with image dimensions ranging from approximately $[85\sim198]\times512\times512$ voxels in $D\times H\times W$ order, where $D$ denotes the slice depth. The in-plane pixel spacing varies between 0.59\,mm and 0.98\,mm, while the slice thickness ranges from 2.50\,mm to 5.00\,mm. 
Following~\cite{liu2023clip}, we perform 5-fold cross-validation, where in each fold 24 volumes are used for training and the remaining 6 volumes for evaluation. \textit{BTCV++} is split into 4,468 slices for training, 473 for validation, and 1,206 for testing. All splits were performed at the \textit{subject level} to avoid data leakage.

For textual supervision, we adopt the multimodal image–text version of MosMedData+ introduced in~\cite{li2023lvit}, where chest CT slices are paired with corresponding clinical textual annotations constructed for medical image segmentation tasks. That dataset comprises 2,729 chest CT slices with pixel-level annotations of COVID-19 infection regions, including ground-glass opacities and consolidations. We use the same data split as in~\cite{li2023lvit},  with 2,183 slices for training, 273 for validation, and 273 for testing.

\subsection{Evaluation Metrics}
For segmentation, we report Dice scores, evaluated independently for each class to assess overall segmentation performance. We additionally report the 95th percentile Hausdorff distance (HD95) for ablation studies.
For detection, we use the mean Average Precision (mAP) over IoU thresholds of 0.1 as the evaluation metric. 
\subsection{Comparison with SoTA methods.}
\noindent\textbf{BTCV++.} 
As shown in Tab.~\ref{tab:seg_det_unified}, our unified model with the Qwen2.5VL-3B backbone achieves the best overall performance for both segmentation and detection. 
Notably, LISA~\cite{lai2024lisa} is less effective on small, thin, or low-contrast structures such as the PSV (+9.42\% Dice improvement) and adrenal glands (right: +7.85\%, left: +3.33\%), suggesting that contrastive vision--language alignment alone is insufficient for fine-grained medical delineation.
These results provide evidence for the need for iterative region-focused analysis, as implemented in our closer-look mechanism for progressive refinement.
For detection, DETR~\cite{carion2020end} is a strong Transformer-based baseline, yet our model surpasses it while simultaneously providing segmentation and interpretable language outputs in a single model.
 \\
\textbf{BTCV~\cite{landman2015miccai}.} Table~\ref{tab:btcv_benchmark_reasoning_style} reports segmentation results on the BTCV benchmark using five-fold cross-validation. 
Overall, all models achieve higher Dice scores compared to \textit{BTCV++} results, indicating that BTCV presents an easier setting with more consistent anatomical appearance, reduced domain shift, and fewer testing variations. 
Although our method operates on 2D slices, it achieves the best overall mean Dice score while maintaining competitive performance across organs compared to 3D methods.
This consistent improvement indicates that integrating task-conditioned visual reasoning with structured perception benefits the standard 3D medical segmentation datasets such as BTCV.
 \\
\textbf{MosMed+~\cite{morozov2020mosmeddata} dataset.} 
As shown in Tab.~\ref{tab:mosmed+}, our method achieves the best Dice and mIoU scores on MosMed+, demonstrating consistent improvements over prior vision–language baselines. 
Among existing baselines, vision–language and contrastive pretraining methods such as CLIP, GLoRIA, and LViT-T achieve competitive performance, highlighting the benefit of semantic alignment for medical image segmentation. 
However, their performance remains limited in accurately delineating fine-grained infection regions, indicating that global vision–language representations alone are insufficient for robust boundary-sensitive segmentation. 
These results further validate that integrating task-conditioned visual reasoning with structured perception enhances segmentation performance across both organ and lesion datasets.

\begin{figure*}[t!]
    \centering
    \includegraphics[width=\linewidth]{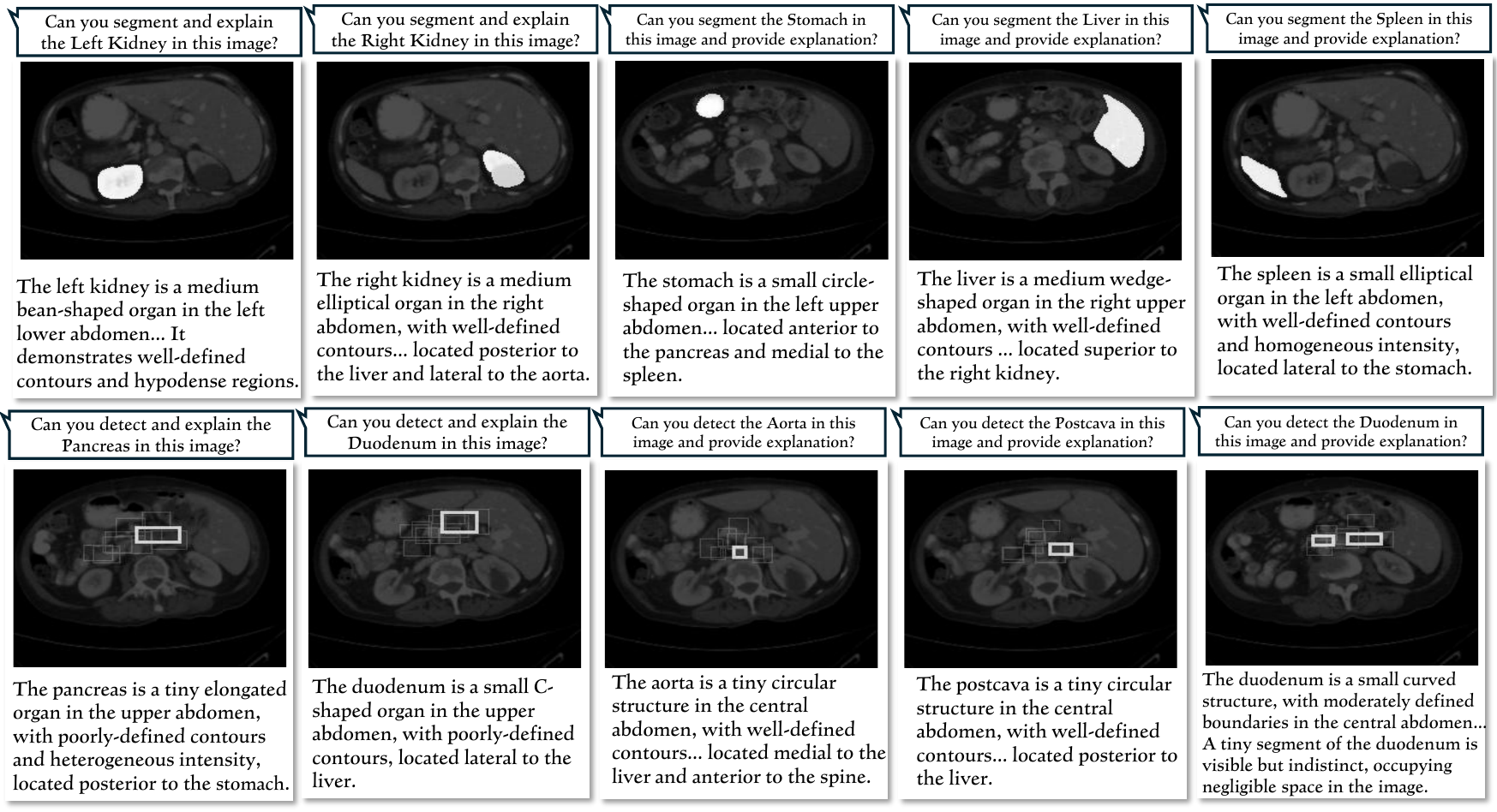}
    \caption{\textbf{Qualitative results} of model outputs on BTCV++. The top row shows segmentation results (together with the linguistic reasoning), while the bottom row presents detection results. Best viewed with zoom-in.}
    \vspace{-15pt}
    \label{fig:qualitive}
\end{figure*}

\subsection{Ablation Study}
\noindent \textbf{Ablation of the closer-look refinement.}
Table~\ref{tab:closerlook_ablation} presents results from an ablation study using BTCV++ to evaluate the contribution of the \emph{closer-look} refinement stage. The results show that including the closer-look mechanism gives consistent improvements across model scales; for example, in Janus1B, the average Dice and NSD increase by approximately 1.68 and 1.30, respectively.
This indicates that re-entering the reasoning--perception loop on region-of-interest patches effectively refines boundary delineation and reduces local segmentation ambiguity. 

\noindent \textbf{Quality of predicted appearance descriptions.} We qualitatively evaluated model outputs to gain insight into its reasoning. Table~\ref{tab:capability_analysis} presents a qualitative comparison on BTCV++ using an \texttt{LLM-as-Judge} evaluation protocol. The results built with a Qwen2.5VL-3B backbone consistently outperforms the Janus1B variant across all evaluated factors. In particular, it improves shape reasoning accuracy by 15.14\% and size understanding accuracy by 5.29\%. Similar gains are observed in contextual and positional understanding, suggesting that a stronger vision–language backbone yields enhanced reasoning capability.\\
\textbf{Ablation of LoRa rank.} As shown in Tab.~\ref{tab:lora}, segmentation performance improves as the LoRA rank increases from 2 to 8, with rank 8 achieving the best results (81.37 Dice and 5.64 HSD). 
Further increasing the rank to 16 slightly reduces performance despite more trainable parameters suggesting diminishing returns. 
This indicates that moderate-rank LoRA provides sufficient adaptation capacity for fine-grained medical segmentation. \\
\textbf{Analysis of Detection Branch.} As shown in Table~\ref{tab:ablation_small_organs}, the detection branch consistently outperforms mask-derived bounding boxes on challenging small and topologically complex organs. In particular, the detection branch improves mAP@0.1 by +10.4 points on the PSV and by +19.2 points on the duodenum compared with mask→box using area-based confidence, corresponding to relative gains of 18.9\% and 28.9\%, respectively. For compact small organs such as the adrenal glands, the detection branch also yields modest yet consistent improvements (+1.35 mAP on RAG and +1.86 mAP on LAG). These results demonstrate that direct learned detection is substantially more robust than post-processed mask-to-box conversion for small and fragmented anatomies.

\section{Visualisation}
\noindent As shown in Fig.~\ref{fig:qualitive}, we visualise qualitative results of models built using BTCV++, where our model demonstrates visually superior results for both detection and segmentation tasks. 

\section{Conclusion}
\noindent In this work, we presented a unified autoregressive framework for CT image interpretation that integrates image analysis and language-based reasoning within a single generative process. By introducing task-routing tokens to coordinate segmentation, detection, and region-focused refinement, the proposed model enables flexible task switching and iterative image-based reasoning, closely reflecting real-world clinical diagnostic imaging workflow. We curated BTCV++ with structured vision–language annotations to support training and evaluation of multimodal grounding and slice-wise reasoning in medical imaging.
Experiments on the BTCV++ benchmark and public datasets demonstrate that the proposed framework achieves state-of-the-art performance in abdominal CT segmentation and detection while providing interpretable, text-based reasoning. A potential limitation of the proposed framework is the increased computational cost introduced by the autoregressive design, reflecting a trade-off for enabling language-guided reasoning and structured outputs. However, these results suggest that coupling language-based reasoning with visual perception offers a promising direction toward more transparent and clinically reliable medical imaging AI systems. 

\bibliographystyle{ieeetr}
\bibliography{reference}
\clearpage
\appendices

\end{document}